\documentclass[utf8]{FrontiersinVancouver} 

\usepackage{url,hyperref,lineno,microtype,subcaption}
\usepackage[onehalfspacing]{setspace}
\usepackage{algorithm}
\usepackage{algpseudocode}
\usepackage{graphicx} 
\usepackage{subcaption} 
\usepackage{caption}
\usepackage{float} 
\usepackage{longtable} 
\usepackage{tabularx}

\usepackage{xcolor}

\hypersetup{
    colorlinks=true,
    linkcolor=cyan,
    filecolor=magenta,      
    urlcolor=cyan,
    citecolor=cyan
    }

\def\keyFont{\fontsize{8}{11}\helveticabold }
\def\firstAuthorLast{Hou {et~al.}}
\def\Authors{Yihao Hou\,$^{1,2}$, Christoph Bert\,$^{1,3,4}$, Ahmed Gomaa\,$^{1,3,4}$, Godehard Lahmer\,$^{1,3,4}$, Daniel H\"ofler\,$^{1,3,4}$, Thomas Weissmann\,$^{1,3,4}$, Raphaela Voigt\,$^{1,3,4}$, Philipp Schubert\,$^{1,3,4}$, Charlotte Schmitter\,$^{1,3,4}$, Alina Depardon\,$^{1,3,4}$, Sabine Semrau\,$^{1,3,4}$, Andreas Maier\,$^{2}$,  Rainer Fietkau\,$^{1,3,4}$, Yixing Huang\,$^{1,3,4,*}$, Florian Putz\,$^{1,3,4}$}
\def\Address{$^{1}$Department of Radiation Oncology, Universit\"atsklinikum Erlangen, Friedrich-Alexander-Universit\"at Erlangen-N\"urnberg, Erlangen, Germany\\
$^{2}$Pattern Recognition Lab, Friedrich-Alexander-Universit\"at Erlangen-N\"urnberg, Erlangen, Germany\\
$^{3}$Comprehensive Cancer Center Erlangen-EMN, Erlangen, Germany\\
$^{4}$Bavarian Cancer Research Center (BZKF), Erlangen, Germany
}
\def\corrAuthor{Yixing Huang}
\def\corrEmail{yixing.yh.huang@fau.de}

\begin{document}
\onecolumn
\firstpage{1}

\title[Fine-Tuning LLaMA-3 for Radiation Oncology]{Fine-Tuning a Local LLaMA-3 Large Language Model for Automated Privacy-Preserving Physician Letter Generation in Radiation Oncology}

\author[\firstAuthorLast ]{\Authors} 
\address{} 
\correspondance{} 

\extraAuth{}

\maketitle

\begin{abstract}

Generating physician letters is a time-consuming task in daily clinical practice. This study investigates local fine-tuning of large language models (LLMs), specifically LLaMA models, for physician letter generation in a privacy-preserving manner within the field of radiation oncology. Our findings demonstrate that base LLaMA models, without fine-tuning, are inadequate for effectively generating physician letters. The QLoRA algorithm provides an efficient method for local intra-institutional fine-tuning of LLMs with limited computational resources (i.e., a single 48 GB GPU workstation within the hospital). The fine-tuned LLM successfully learns radiation oncology-specific information and generates physician letters in an institution-specific style. ROUGE scores of the generated summary reports highlight the superiority of the 8B LLaMA-3 model over the 13B LLaMA-2 model. Further multidimensional physician evaluations of 10 cases reveal that, although the fine-tuned LLaMA-3 model has limited capacity to generate content beyond the provided input data, it successfully generates salutations, diagnoses and treatment histories, recommendations for further treatment, and planned schedules. Overall, clinical benefit was rated highly by the clinical experts (average score of 3.44 on a 4-point scale). With careful physician review and correction, automated LLM-based physician letter generation has significant practical value.

\tiny
 \keyFont{ \section{Keywords:} Radiation Oncology, ChatGPT, Data Privacy, Parameter-Efficient Fine-Tuning, LLaMA, Fine-Tuning, Physician Letter} 
\end{abstract}

\section{Introduction}

Recently, advancements in neural network architectures \cite{huang2024principles}, such as Transformers \cite{vaswani2017attention}, and effective training strategies, including supervised fine-tuning (SFT) \cite{ziegler2019fine} and reinforcement learning with human feedback (RLHF) \cite{christiano2017deep}, have significantly enhanced the capabilities of large language models (LLMs). Coupled with the increasing availability of computational resources and extensive training data, these developments have led to the release of several prominent LLMs, such as ChatGPT \cite{brown2020language,thapa2023chatgpt}, Gemini \cite{islam2024gemini}, LLaMA \cite{touvron2023llama2}, and PaLM \cite{singhal2022large}. These models have revolutionized diverse domains, including medicine \cite{singhal2023large}, by bringing transformative impacts on various applications.


In addition to their general knowledge, LLMs have demonstrated a certain level of specialized medical expertise including the field of radiation oncology. The general capabilities and limitations of GPT-4 within radiation oncology have been discussed extensively \cite{putz2024exploring}. The performance of LLMs has been benchmarked using the standard ACR Radiation Oncology In-Training (TXIT) exam \cite{huang2023benchmarking}, custom radiation oncology physics questions \cite{holmes2023evaluating}, patient care questions \cite{yalamanchili2024quality}, and other general multiple-choice questions in radiation oncology \cite{dennstadt2024exploring}. Additionally, the efficacy of GPT-4 has been demonstrated in handling real, complex cases from the Red Journal Gray Zone \cite{huang2023benchmarking}. LLMs have shown promise in various radiation oncology tasks, such as medical education through interactive teaching \cite{ebrahimi2023chatgpt}, facilitating research \cite{guckenberger2023potential}, standardizing radiotherapy structure names \cite{syed2020integrated}, obtaining informed patient consents \cite{moll2024role}, exploring personalized treatment pathways \cite{lin2024natural}, and automatically extracting radiation therapy events \cite{bitterman2023end,choi2023developing}. However, since LLMs can generate convincing but false responses, there is a risk of inexperienced users overtrusting these AI-generated outputs \cite{guckenberger2023potential}. To mitigate such hallucination problems, a new method called ReAct (Reason + Act) has been proposed for treatment decision support, which constrains GPT-4's responses based on given treatment guidelines through in-context learning \cite{putz2024exploring}.

Automation in the healthcare sector by LLMs could have great importance to maintain patient care into the future \cite{janssen2024survey}, while enabling cost-efficient healthcare systems that offer a high standard of care. 
Because of the dramatic demographic changes in most western countries, an increase in patients requiring health care services is projected to meet a shrinking supply of healthcare workers in the coming years \cite{jones2024healthcare}. Already by 2030, a shortage of 1.2 million registered nurses and 121,900 physicians is expected for the US \cite{markit2017complexities,congressional20142014}, while a deficit of 488,000 health care workers has been forecasted for the UK \cite{congressional20142014}. Partial automation of time-consuming simple or bureaucratic tasks with LLMs in the health care sector could make health care systems more efficient and mitigate the expected demographic impact. As a shortfall in physicians has been shown to increase patient mortality \cite{HealthFunding2021}, LLM-automation of simple tasks like physician letter generation, could even positively affect clinical outcomes by freeing up physician resources for the tasks where they are needed the most.

Despite the promise of LLMs in various radiation oncology applications and the broader field of medicine, data privacy remains a pressing concern, particularly under regulations such as the EU Medical Device Regulation \cite{beckers2021eu} and the EU General Data Protection Regulation (GDPR) for health data. Most LLMs, including GPT-4, are proprietary AI models. Their use in clinical settings requires data sharing to external AI hosting service providers, raising significant security and privacy issues for patient data. For instance, although ChatGPT users can disable historical chat logs, conversation data is retained for 30 days to monitor data misuse according to OpenAI documents \cite{openai_blog_chatgpt_data_management}. Furthermore, OpenAI has faced criticism for allegedly using private or copyrighted data to train GPT-4 without obtaining necessary consent agreements \cite{khowaja2024chatgpt}.
To address data privacy concerns, open-source LLMs such as LLaMA \cite{touvron2023llama2} have emerged, which can be deployed locally within hospitals. Local training and inference of LLMs within a hospital's local IT infrastructure is very promising, as it eliminates the risk for data sharing, maximizes patient data safety and minimizes regulatory issues. Fine-tuned LLaMA-2 models have been reported to achieve performance comparable to proprietary counterparts like GPT-3.5 \cite{nievas2024distilling}. Examples of such fine-tuned LLaMA models, including ChatDoctor \cite{li2023chatdoctor} and HuaTuo \cite{wang2023huatuo}, have demonstrated promising performance in clinical knowledge applications. In April 2024, MetaAI released LLaMA-3 \cite{dubey2024llama}, which is announced to offer performance comparable to GPT-4.
In this work, we aim to fine-tune and evaluate LLaMA-3 as a local LLM for the task of generating physician letters in the field of radiation oncology, illustrating how LLM technology can be leveraged in clinical practice by local deployment within hospitals.

\section{Methodology}

\subsection{Dataset Construction}

In this study, two types of texts were generated using fine-tuned LLMs: patient case summary reports and physician letters. Both types of texts provide an essential overview of patients' situations, aiding other physicians, healthcare providers, and patients in understanding and communicating the most important medical characteristics of a patient case. Summary reports are commonly used in tumor board and ward round presentations as well as within electronic health records, while physician letters inform patients or other medical departments about diagnoses, medical history, and treatment plans. Manually writing these letters often is a time-consuming and tedious task for physicians, which to a large part may involve rearranging textual information that is represented elsewhere, e.g. in previous medical documents. Therefore, the automatic generation of such letters using a localized LLM holds significant clinical value.

A set of physician letters were collected from the Department of Radiation Oncology at University Hospital Erlangen, Germany, spanning from 2010 to 2023. For the generation of summary letters, 560 cases with comprehensive diagnosis and treatment records were extracted and formatted in a question-and-answer style for fine-tuning. In the input, all the private information of the patient, e.g. patient name, birthday, and patient ID, was removed, and only essential diagnosis and treatment information was kept. The summary report generation task was a trial experiment for us to determine optimal fine-tuning parameters for the physician letter generation task. For physician letter generation, 14,479 letters were used for fine-tuning, where all the information including patient- and physician-specific private information was kept. Ten cases for each task that were completely independent from the training cases were reserved for testing. The model's input was the tabular data of the original physician letter head, which included the date of the document creation, the physician author of the letter, the patient demographic information, diagnoses and medical history, the planned treatment as well as the recipient of the letter (Fig.\,\ref{Fig:InputExample2}). For practical use at our institution, this information can be simply copied from other sources, significantly enhancing efficiency for practical deployment. The model was fine-tuned to predict the written section of the original physician letter beginning with the salutation and ending with the physician signatures (Fig.\,\ref{Fig:OutputExample2}). Given that German is the official language at our hospital and the pretrained LLMs have the capability to understand German, all input information was in German. The summary reports were generated in English, while the physician letters were generated in German to allow for realistic evaluation by the assessing physician raters as well as for actual clinical deployment. For the purpose of this manuscript, all letters were translated into English to facilitate understanding by the international community. All letter excerpts shown in this manuscript were fully anonymized, which included shifting of dates by an arbitrary interval, while preserving the relative time intervals within a physician letter as well as the aspects relevant to the results and the discussion.

\subsection{Model Fine-Tuning}
\subsubsection{Base Models}
The LLaMA-2 \cite{touvron2023llama2} and LLaMA-3 \cite{dubey2024llama} models, released by Meta on July 18, 2023, and April 18, 2024, respectively, were used as pretrained base models. These models can be fine-tuned locally within an institution, ensuring data privacy during both the fine-tuning and final deployment phase. The LLaMA-2 family includes pretrained models with parameter sizes of 7B, 13B, 34B, and 70B, where larger parameter sizes indicate higher generation capabilities but also require significantly more computational resources. The LLaMA-3 family offers models in two sizes: 8B and 70B. Each pretrained model has a corresponding instruction fine-tuned version for dialog-related tasks (e.g., LLaMA-3-8B-Instruct) as well as a general, non-instruction fine-tuned version for text completion tasks (e.g., LLaMA-3-8B). For the tasks of patient cases summarisation and physician letter generation in this work, the general, non-conversational LLaMA-3 model variants were directly fine-tuned for their respective downstream tasks. Due to limited computational resources available in a hospital setting, the 13B LLaMA-2 model and the 8B LLaMA-3 model were utilized for further fine-tuning. 

\subsubsection{Low-Rank Adaptation of LLMs}
Due to the large number of parameters in LLMs, it is inefficient to fine-tune all the parameters. Therefore, parameter-efficient fine-tuning (PEFT) techniques \cite{peft,li2021prefix,liu2023gpt} are preferred, which keep the parameters of pretrained LLMs frozen and only need to train a few parameters added for a specific down-stream task. Some PEFT methods \cite{peft,rebuffi2017learning,lin2020exploring} apply adapter modules for fine-tuning, which achieve fine-tuning effectively, but lead to latency in inference due to the lack of parallelism at the additional adapters. Prompt fine-tuning methods \cite{li2021prefix,liu2023gpt} are challenging to search for optimal prompts and typically lead to reduced performance due to the reduced token size available for down-stream tasks.  Since LLMs are typically overly parameterized and their performance relies on certain intrinsic low dimensions \cite{aghajanyan2021intrinsic,li2022low}, low-rank adaptation (LoRA) \cite{LORA} of LLMs has emerged as the most widely adopted method of the PEFT family. 

The fundamental idea of LoRA is illustrated in Algorithm\,\ref{alg:algorithm2}. When a pretrained LLM is denoted by a high-dimensional matrix $\boldsymbol{W}_0\in \mathbb{R}^{d\times k}$ with large dimension sizes $d$ and $k$, its fine-tuned version is denoted by $\mathbf{W}_1 \in \mathbb{R}^{d\times k}$, which can be decomposed as $\mathbf{W}_1 = \mathbf{W}_0 + \Delta \mathbf{W}$. According to the low-rank assumption, the difference $\Delta \mathbf{W}$ can be represented by the multiplication of two matrices $\mathbf{A} \in \mathbb{R}^{r\times k}$ and $\mathbf{B}\in \mathbb{R}^{d\times r}$, i.e., $\Delta \mathbf{W} = \mathbf{B}\mathbf{A}$, where the dimension/rank $r$ is much smaller than $d$ and $k$.
Because of the low rank design,  LoRA is much more efficient in computation than other PEFT methods.  Moreover, as the additional parameters of $\mathbf{A}$ and $\mathbf{B}$ are added in parallel to the pretrained LLM parameters $\boldsymbol{W}_0$, the latency problem in inference is avoided.

\begin{algorithm}[t]
\caption{LoRA: Low-Rank Adaptation for LLMs \cite{LORA}}
\label{alg:algorithm2}
\begin{algorithmic}
\Function{LoRA}{$\mathbf{W}_0, \mathbf{A}, \mathbf{B}, \mathbf{x}$}
    \State $\text{Frozen Input: Pre-trained weight matrix } \mathbf{W}_0 \in \mathbb{R}^{d \times k}$
    \State $\text{Trainable additional parameters for fine-tuning: Low-rank matrices } \mathbf{A} \in \mathbb{R}^{r \times k}, \mathbf{B} \in \mathbb{R}^{d \times r}$
    \State $\text{Input: Input representation } \mathbf{x} \in \mathbb{R}^{d}$
    \State $\text{Output: Adapted output representation } \mathbf{y} \in \mathbb{R}^{k}$
    \State
    \State $\mathbf{y} \gets \mathbf{W}_0\mathbf{x} + (\mathbf{B}\mathbf{A})\mathbf{x}$ \Comment{Apply LoRA reparametrization}
    \State \Return $\mathbf{y}$
\EndFunction
\end{algorithmic}
\end{algorithm}

A key objective of this work is to develop a standardized workflow that enables small-scale medical institutions to fine-tune their own LLMs using local, private medical data. Reducing training costs and computational expenses is therefore highly significant. In this context, the quantized LoRA (QLoRA) algorithm \cite{dettmers2024qlora} provides a more memory- and computation-efficient fine-tuning solution compared to standard LoRA. QLoRA utilizes quantization techniques to convert conventional 16-bit pre-trained LLMs into 8-bit or 4-bit low-precision models, maintaining performance without significant degradation \cite{dettmers2024qlora}. Additionally, QLoRA introduces paged optimizers \cite{dettmers2024qlora}, which address the out-of-memory issue caused by memory spikes during training. This is achieved by temporarily offloading optimizer states from the GPU to the CPU memory, allowing the GPU to handle immediate high memory demands without crashing. Once memory usage stabilizes, the state is transferred back to the GPU. This approach significantly enhances the feasibility of training large models in resource-constrained environments.

\subsection{Experimental Setup}
\subsubsection{Training details}

The base LLaMA models (the 13B LLaMA-2 model and the 8B LLaMA-3 model) were fine-tuned with QLoRA using two NVIDIA A6000 GPUs (48 GB memory). A max length of 1500 and 2000 tokens, respectively, was set for the input sequences fed to the LLaMA models for the patient case summarisation and physician letter generation tasks. The LoRA rank $r$ was set to 32 and a scaling factor of 64 to increase the contribution of low-rank adaptions. The dropout rate for LoRA was set to 0.05. The target weight matrices in LLaMA, which LoRA was applied to, include q\_proj, k\_proj, v\_proj, o\_proj, gate\_proj, up\_proj, down\_proj, and lm\_head. The 8-bit paged Adamw optimizer was used with a learning rate of $1 \times 10^{-5}$. The batch size for each GPU was 2 and parallel training using two GPUs were enabled. Gradient accumulation steps were set to 2 to allow for larger effective batch sizes without requiring more memory. 500 total iteration steps were applied for the summary report. For the physician letter generation task, 15,000 iteration steps were applied, which took around 58 hours. Around 30 GB (\%60) and 23 GB (48\%) of GPU memory were used for fine-tuning the 13B LLaMA-2 and 8B LLaMA-3 models, respectively.

\subsubsection{Evaluation metrics}
The ROUGE scores \cite{lin2004rouge} and a multidimensional expert rating by five physicians were used to evaluate the performance of the LLMs.

\textbf{ROUGE Scores:}
ROUGE \cite{lin2004rouge} is short for Recall-Oriented Understudy for Gisting Evaluation, which is a common metric in the field of natural language processing (NLP). It compares a model's text output with a reference text, e.g., a human generated text for the same input, to evaluate the similarity. ROUGE scores can range from 0 to 1, with higher values indicating a greater alignment between the model output and its corresponding reference text.

ROUGE scores have different variants, commonly known as ROUGE-N (including ROUGE-1, ROUGE-2 and ROUGE-L), which is computed based on N-grams. An N-gram is a term of N words. For example,  a reference sentence ``I love machine learning" is divided to a list of [``I", ``love", ``machine", ``learning"] for 1-grams, and  a list of [``I love", ``love machine", ``machine learning"] for 2-grams. Correspondingly, an output candidate sentence ``I like machine learning very much" is divided to [``I", ``like", ``machine", ``learning", ``very", ``much"] for 1-grams and [``I like", ``like machine", ``machine learning", ``learning very", ``very much"] for 2-grams, respectively. 
With such N-grams, the recall, precision, and F1 measures of ROUGE-N metrics can be computed. Recall is defined as the overlapping number of N-grams divided by the number of N-grams in the reference, e.g., recall of ROUGE-1 = 3/4 for the given example; precision is defined as the overlapping number of N-grams divided by the number of N-grams in the candidate sentence, e.g., precision of ROUGE-1 = 3/6 for the given example. The F1 measure is defined as F1 = 2 * recall * precision / (recall + precision), e.g., F1 measure of ROUGE-1 = 0.6. Note that in the example ``love" and ``like" have a similar semantic meaning, but are considered as different words in ROUGE scores.

\textbf{Expert rating:} The ROUGE scores provide a quantitative analysis of the similarity between reference and LLM-generated physician letters. However, ROUGE scores have a lot of limitation in evaluating the medical context. Therefore, the generated physician letters were further evaluated on a 4-point scale across multiple dimensions by 5 physicians: correctness, comprehensiveness, clinic-specific style, and practicality. The scores for different dimensions are defined as the following:

\begin{itemize}
\item Correctness:\\ 
    Score 1 - Serious errors, risk for incorrect clinical decisions \\ 
    Score 2 - Relevant errors, without clinical impact \\ 
    Score 3 - Minor inaccuracies, irrelevant to the patient case \\ 
    Score 4 - The letter contains no errors 
\item Comprehensiveness (need for adjustments): \\
    Score 1 - The letter is so incomplete that it is faster to rewrite the letter \\ 
    Score 2 - The letter needs major adjustments $>$ 1 min \\ 
    Score 3 - The letter needs minor adjustments $\leq$ 1 min \\ 
    Score 4 - The letter is complete and does not require any adjustments 
\item Clinic/institute specific content and style:\\
  	Score 1 - No clinic-specific content or adaptation to the local letter style \\
  	Score 2 - Very little clinic-specific content or adaptation to the local style \\
  	Score 3 - The letter contains significant clinically specific content
or adaptations to the local style \\
	Score 4 - The letter completely reflects the style of a local letter

\item Benefit in practice (practicality):\\ 
    Score 1 - No use for letter writing \\ 
    Score 2 - Small benefit for letter writing\\ 
    Score 3 - Moderate benefit for letter writing  \\ 
    Score 4 - Great benefit for letter writing
\end{itemize}

For correctness, a note was added: Please do not evaluate dates that are not included in the input data, these are estimated by the AI based on the dates within the input data.

\begin{figure}[t]
    \centering
    \includegraphics[width=0.8\linewidth]{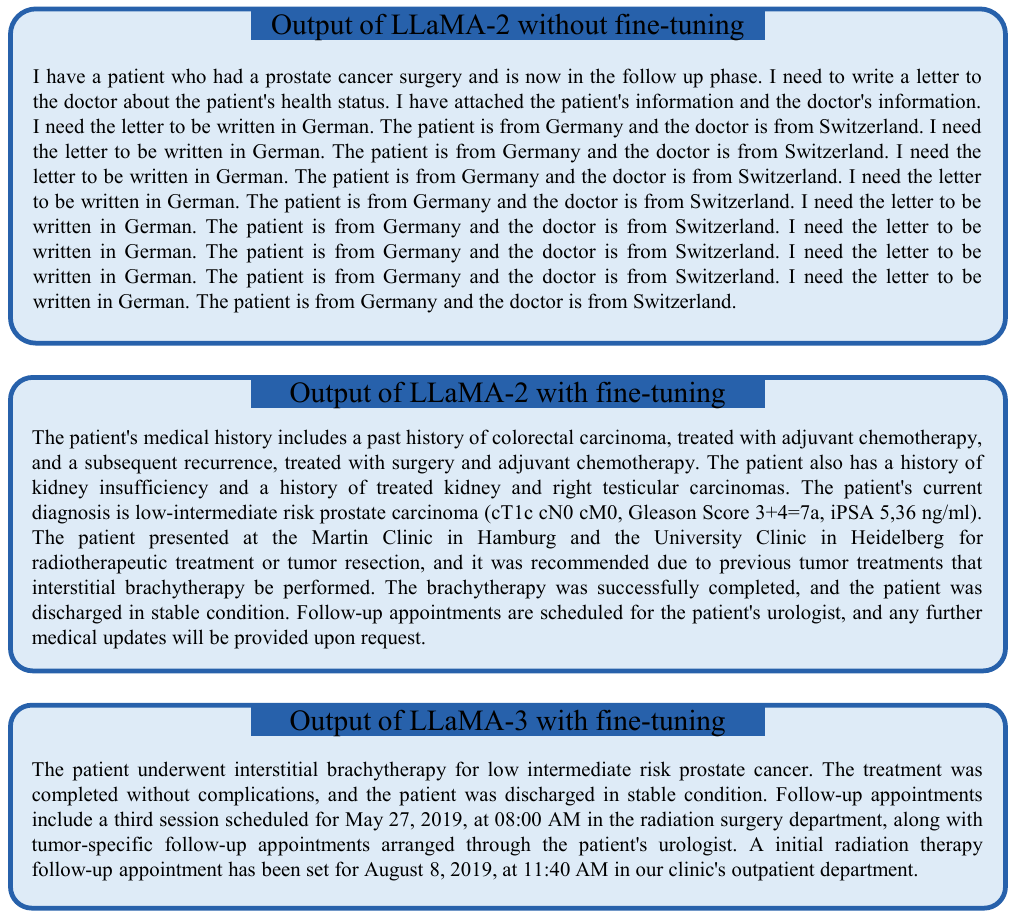}
    \caption{The outputs of the locally fine-tuned LLaMA-2 (center) and LLaMA-3 (bottom) models compared to the baseline LLaMA-2 model in an exemplary case for the task of patient case summarisation. }
    \label{Fig:OutputExample}
\end{figure}

\begin{figure}[t]
  \centering
  \begin{minipage}[t]{0.75\textwidth}
  \includegraphics[width=\linewidth]{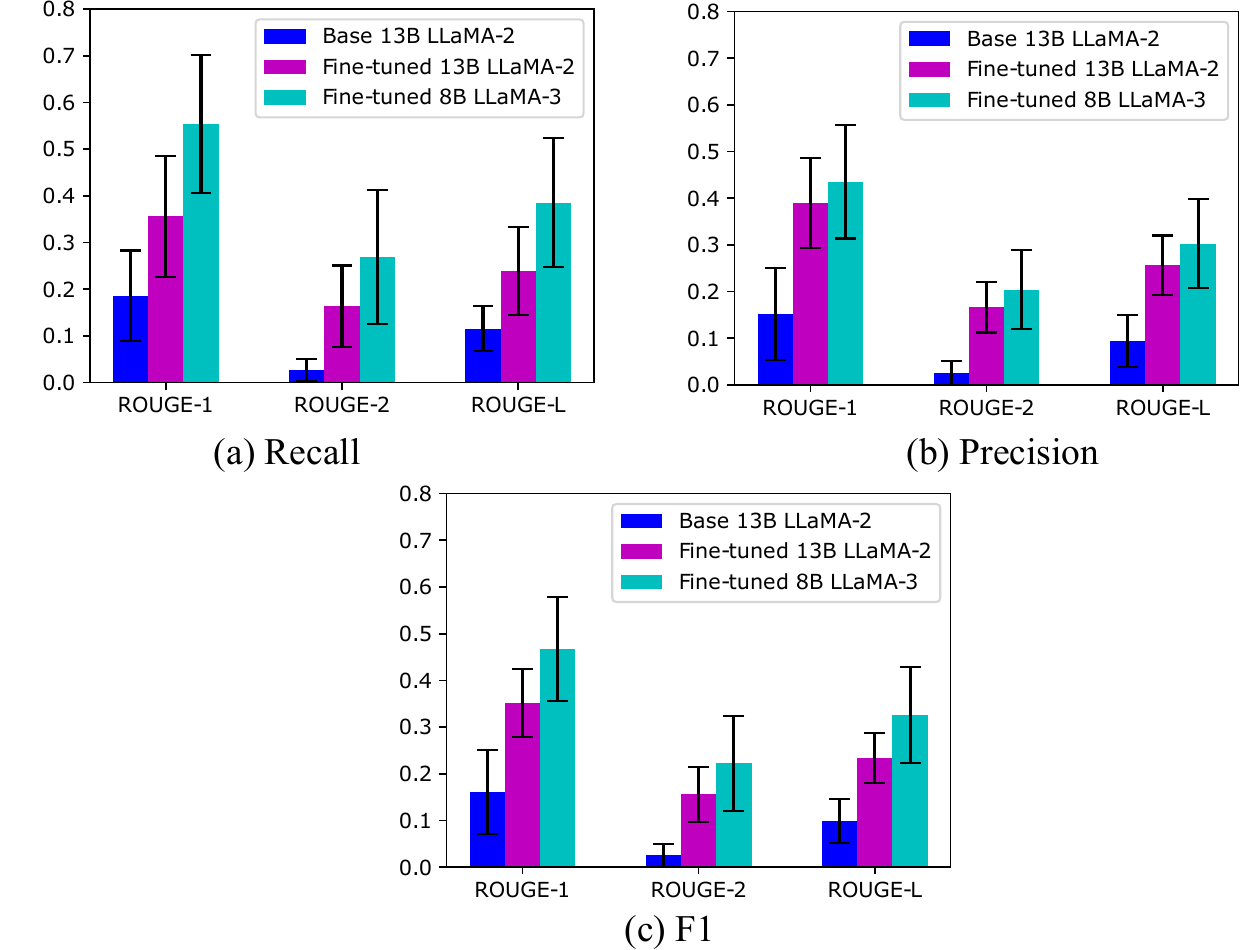}
  \end{minipage}
  \caption{The ROUGE scores of LLaMA models for the task of patient case summarisation with and without local fine-tuning on institutional data. The error bars indicate standard deviations.}
  \label{Fig:13BModelROUGE}
\end{figure}

%

%

\section{Results}
\subsection{Summary report generation task}

The input data of an exemplary case is displayed in Supplementary Fig. 1. The input document for the patient case summarisation task included the patient's primary diagnoses, secondary diagnoses, tumor-specific history, clinical course and planned follow-up procedures.  Without fine-tuning, the 13B LLaMA-2 model generated some texts irrelevant to the input case, as displayed in Fig.\,\ref{Fig:OutputExample}. In contrast, the fine-tuned LLaMA-2 and LLaMA-3 models both provided a relevant summary of the patient case despite some inaccuracies, as displayed in Fig.\,\ref{Fig:OutputExample}.

 The ROUGE scores for 10 patient case summaries generated by the LLaMA-2 and LLaMA-3 models are displayed in Fig.\,\ref{Fig:13BModelROUGE}. The F1 measures of ROUGE-1, ROUGE-2, and ROUGE-L were 0.161, 0.025, and 0.099 for the 13B LLaMA-2 model without fine-tuning, respectively. After fine-tuning LLaMA-2, they were improved to 0.352, 0.156 and 0.234 with statistical significance (p$\leq$0.01 paired t-test, Fig.\,\ref{Fig:13BModelROUGE}(c)) for ROUGE-1, ROUGE-2, and ROUGE-L, respectively. This highlights the benefit of fine-tuning. Interestingly, compared with the fine-tuned 13B LLaMA-2 model, the fine-tuned 8B LLaMA-3 model further improved the ROUGE scores, despite its lower number of model parameters.

\subsection{Physician letter generation}

\begin{figure}[b]
\includegraphics[width=\linewidth]{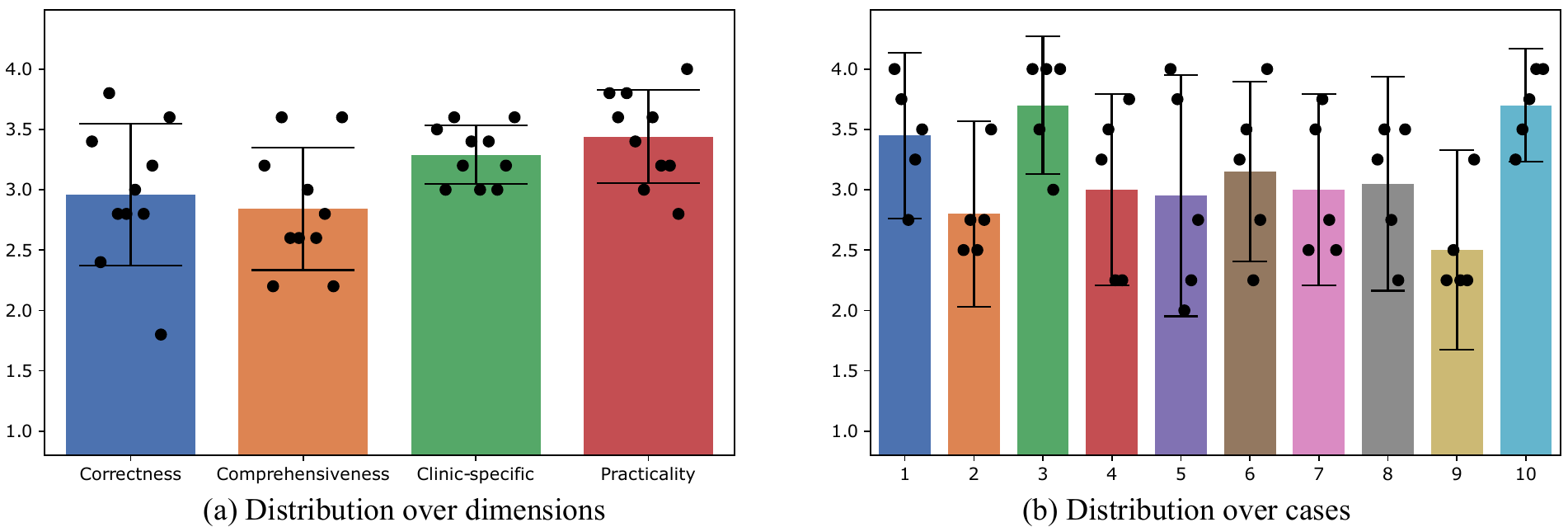}
 \caption{The distribution of average physician rating scores for the physician letters automatically generated by the locally fine-tuned 8B LLaMA-3 model. The error bars indicate standard deviations.}
    \label{Fig:PhysicalRating2}
\end{figure}

\begin{figure}[t]
    \centering
    \includegraphics[width=0.8\linewidth]{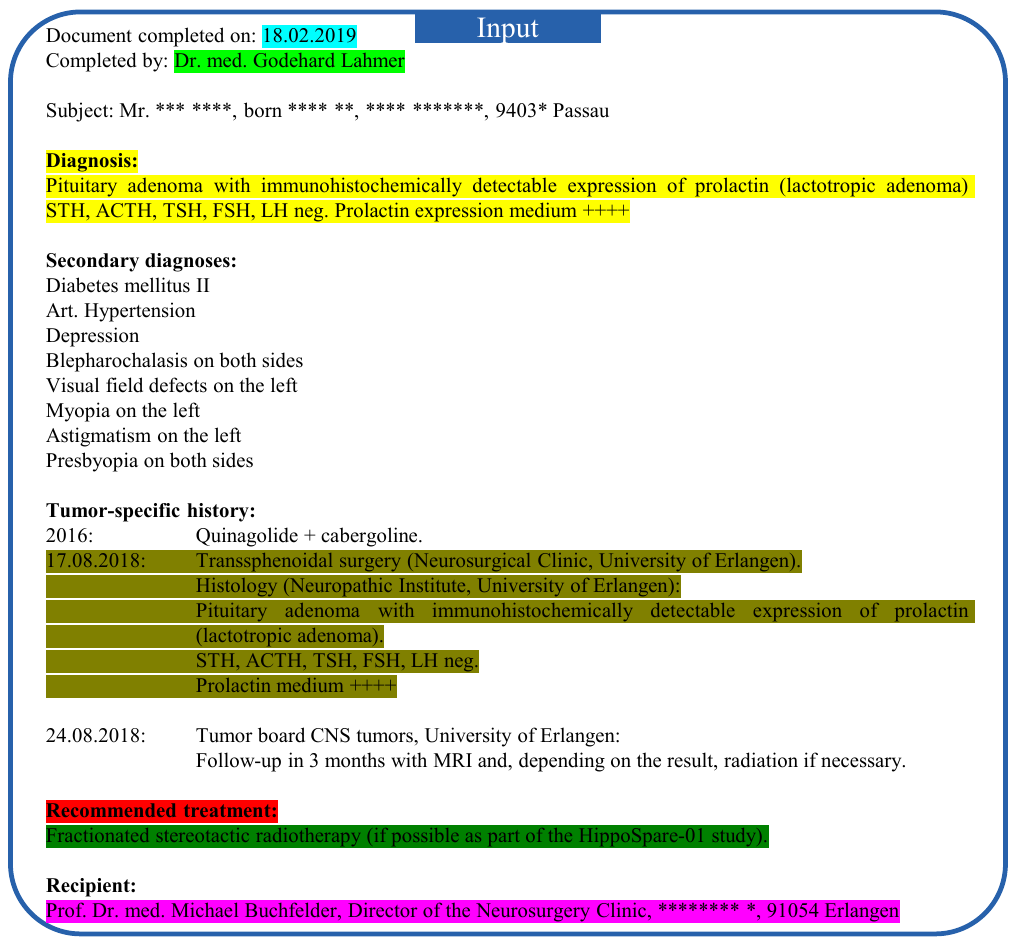}
    \caption{Input medical data of Case $\#$1 for the physician letter generation task. Note that some keywords are highlighted in bold by the authors for better visualization, but the content was provided in plain text to the LLM. Certain private information is anonymized with the symbol $\ast$. Different segments of the patient input information in regard to the model output (Fig.\,\ref{Fig:OutputExample2}) are highlighted by different colors.}
    \label{Fig:InputExample2}
\end{figure}

Due to the superior performance of the fine-tuned 8B LLaMA-3 model, it was selected for the subsequent automatic physician letter generation task. The input data for the automated physician letter generation task included the data from the original letter head including the date and physician author of the letter, the recipients of the letter, the patient's demographic information, diagnoses, as well as the medical history with information on planned or recommended future procedures in tabular form (Fig.\,\ref{Fig:InputExample2}).
Ten physician letters automatically created by the locally fine-tuned 8B LLama-3 model were evaluated by 5 physicians across four dimensions.
The distributions of physician rating scores over evaluation dimensions and cases are displayed in Fig.\,\ref{Fig:PhysicalRating2}(a) and (b), respectively. The generated physician letters got average scores of 2.96, 2.84, 3.29, and 3.44 over correctness, comprehensiveness, clinic-specific style, and practicality, indicating the decent performance of the locally fine-tuned LLM. Among all the cases, Case $\#$3 and Case $\#$10 achieved high average scores of 3.7 over all the evaluation dimensions, whereas Case $\#$9 got the lowest average score of 2.5.

\begin{figure}[t]
    \centering
    \includegraphics[width=0.8\linewidth]{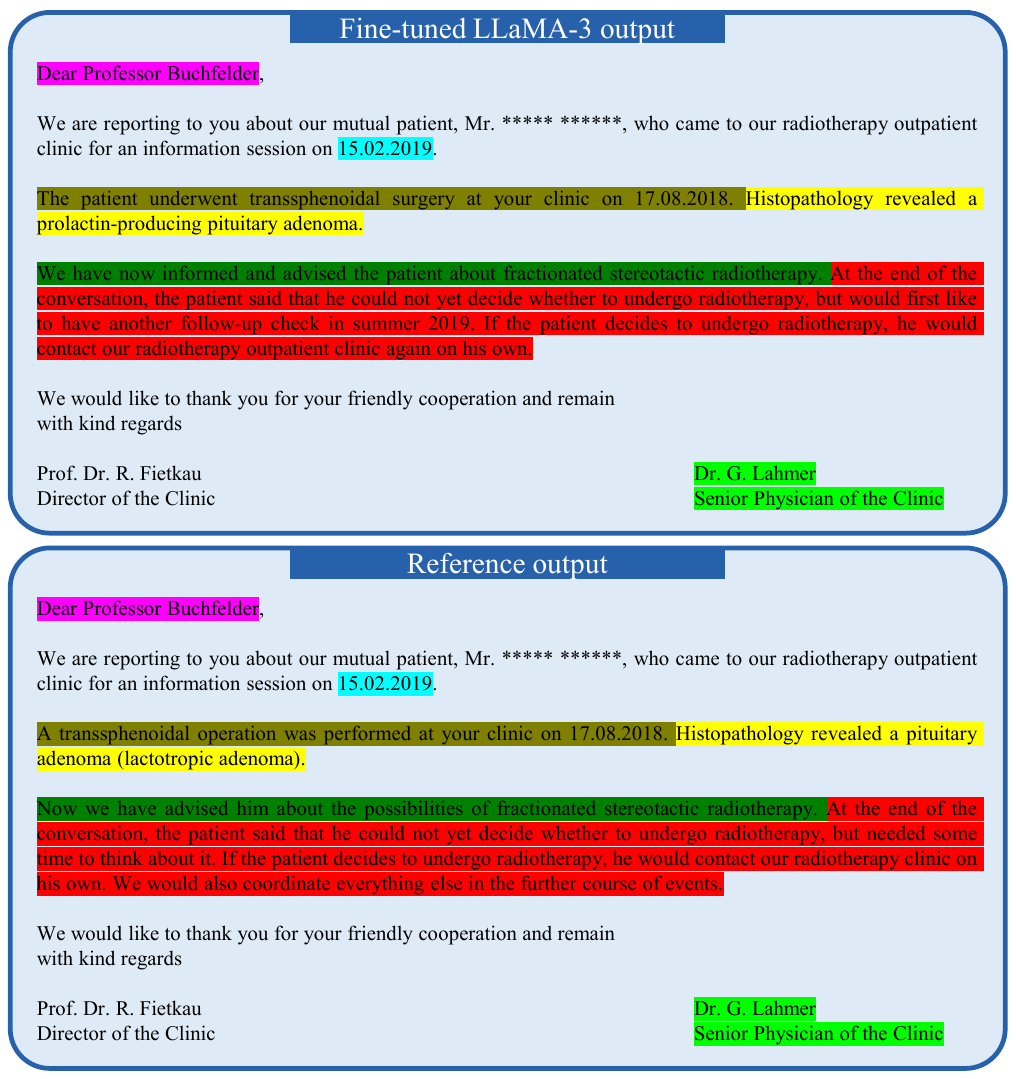}
    \caption{Fine-tuned 8B LLaMA-3 model output of Case $\#$1 for the physician letter generation and its corresponding reference output. The patient name is anonymized with the symbol $\ast$. The highlighted text segments correspond to the information in the input data (Fig.\,\ref{Fig:InputExample2}) highlighted with the same color.  }
    \label{Fig:OutputExample2}
\end{figure}

The input data of Case $\#$1 is displayed in Fig.\,\ref{Fig:InputExample2}. The fine-tuned LLaMA-3 output and the reference output (original letter) are displayed in Fig.\,\ref{Fig:OutputExample2}, where the highlighted text segments correspond to the text sections with matching colors in the input data (Fig.\,\ref{Fig:InputExample2}). For the illustrative example of Case $\#$1, the output of the fine-tuned LLaMA-3 model is correct in the following aspects:

\begin{itemize}
\item Salutations: The model correctly recognized the salutations of the recipient, the director of the clinic, and the physician of the clinic. Note that the information of the director (i.e., Prof. Dr. R. Fietkau) and the senior physician title of the letter author was not directly provided in the input data. However, the model has learned such information from local fine-tuning based on the large number of institution-specific training letters.

\item Consulting date: The date of the physician letter was 18.02.2019. In the generated letter, instead of copying this date, the model chose an earlier date for the date of the actual consultation, which is accurate since physician letters in the training and test data had usually been written one day after the consultation. However, rather than selecting 17.02.2019 (one day earlier), the model opted for 15.02.2019. In this specific case it appears that the model correctly inferred that 17.02.2019 was a Sunday and thus selected the last workday, which was Friday, 15.02.2019.

\item Diagnosis and treatment history: In the given case, the model correctly recognizes the primary diagnosis (prolactin-producing pituitary adenoma) and the past treatment of transsphenoidal surgery in the clinic of the recipient.

\item Recommended treatment: The model also correctly summarized the recommended treatment method, which is fractionated stereotactic radiotherapy.

\item Correct prediction that the patient needs further time for consideration: For Case $\#$1, the keyword ``recommended treatment" is present, prompting the model to predict that the patient needs more time to consider the recommended treatment methods. In other cases, the keyword ``recommended treatment" is replaced with the keyword ``planned treatment" (see Supplementary Document). Interestingly, we found in systematic experiments that the single keyword ``recommended treatment" vs. ``planned treatment" in the input text seems to determine the general content of the output text for letters from the physician author Dr. G. Lahmer. When ``planned treatment" is used, the model predicts that the patient has agreed to the recommended treatment methods and summarizes the specific scheduled radiotherapy planning procedures for the patient. In contrast, if the keyword "recommended treatment" is used instead, the model always predicts that the patient requires more time for consideration. We further observed that this switch-like effect of the single keyword ``recommended treatment" vs. ``planned treatment" on the LLM output is mostly specific to Dr. G. Lahmer indicating that the LLM was able to learn the writing habits of different physicians.

\end{itemize}

The fine-tuned LLaMA3 model had decent performance for Case $\#$1. However, the scores of certain cases were lower, e.g., Case $\#$2 and Case $\#$9. In the input of Case $\#$2, a recommendation of ``interstitial brachytherapy alone, e.g. as permanent brachytherapy with iodine seeds" was provided. With such input information, the fine-tuned model predicted the potential treatment approaches of surgery and radiotherapy, which is correct in general. However, in the original letter, different radiotherapy treatment approaches were discussed in more detail, which included external-beam radiotherapy (EBRT), combined EBRT with brachytherapy boost, permanent brachytherapy with iodine seeds alone, temporary brachytherapy with iridium-192, pulsed dose rate (PDR) brachytherapy, and high dose rate (HDR) brachytherapy. Moreover, the original letter specified the patient's preferred treatment time in spring 2022. Because of such missing details, the output achieved a mean score of 2.8, which is relatively low.

\begin{table}[h!]
\begin{small}
\centering
\begin{tabular}{|p{2cm}|p{14.4cm}|}
\hline
\center{\textbf{Input}} & ...

09/2014: HIFU hemiablation on the left.

02/2016: HIFU hemiablation on the left.

...

Recommendation: Interstitial brachytherapy alone, e.g. as permanent brachytherapy with iodine seeds.

...\\
\hline

\center{\textbf{Output}} &...

 In the case of prostate cancer and history after two HIFU treatments of the left prostate, imaging now shows a high suspicion of a local recurrence on the left posterolateral side. We spoke to the patient about potential treatment methods (\textcolor{red}{surgery or radiotherapy}).
In particular, we explained the options for radiotherapy to him and recommended interstitial brachytherapy alone. At the end of the conversation, the patient said that he now needed some time to think about it. He would contact us himself if he wanted radiotherapy.

...\\
\hline
\center{\textbf{Reference}} & ...

In the case of low-risk prostate cancer and history after 2x HIFU, we have now spoken to the patient with a persistent PSA value of approx. 7 ng/ml about potential radiotherapy methods (\textcolor{red}{percutaneous radiotherapy, combined percutaneous radiotherapy with brachytherapy boost, permanent brachytherapy with iodine seeds alone, temporary brachytherapy with iridium-192, PDR procedure, HDR procedure}). The patient is leaning towards treatment with iodine seeds. At the end of the conversation, he said that if he wanted radiotherapy, he would like to have it done in \textcolor{red}{spring 2022} and would contact him himself to coordinate further appointments.

...\\
\hline
\end{tabular}
\end{small}
\caption{The LLM input, output, and reference from the original physician letter for Case $\#$2. The texts in red highlight the major differences between the fine-tuned LLaMA-3's output and the original physician letter.}
\label{table:Case2}
\end{table}

\begin{table}[t]
\begin{small}
\centering
\begin{tabular}{|p{1.7cm}|p{14.8cm}|}
\hline
\center{\textbf{Input}} & 
\textbf{Diagnoses:}
Rectal carcinoma lower third cT3 cN+ cM0

\textbf{Secondary diagnoses:}
Bland liver cysts

\textbf{Tumor-specific anamnesis:}

01/2020: Blood in the stool.

Colonoscopic diagnosis with removal of multiple polyps, including the distal third of the rectum with exophytic growth.

Histopathological adenoma parts with high-grade suspected carcinoma.

Diagnostic staging completion MRI/CT/endosonography: Suspected deep-seated rectal carcinoma, cT3 cN+ cM0.

\smallskip

01/21/2020: Interdisciplinary tumor conference consensus (University Hospital Regensburg):
Neoadjuvant radiochemotherapy for curative surgery and additional appendectomy (polyp residue).

\smallskip 

02-03/2020: Definitive radiochemotherapy (radiotherapy Regensburg):
1.8 Gy to 50.4 Gy in the area of the rectum and corresponding lymph drainage area.

Chemotherapy:
2 courses of 5-Fu.

\smallskip

04/2020: Patient refuses surgical resection.
Change of concept from neoadjuvant to definitive, corresponding local dose escalation in the area of the primary tumor up to 60.4 Gy + additional 2 courses of FOLFOX.

\smallskip 

06/2020: Rectoscopic residual tumor cT2 cN0.

Planned: Evaluation of the possibility of interstitial brachytherapy.\\
\hline

\center{\textbf{Output}} &...

 The patient was definitively irradiated for a rectal carcinoma using neoadjuvant chemoradiotherapy. The patient refused the recommended surgical resection, so that chemoradiotherapy was considered definitive and the dose was increased locally accordingly.
Ms. ***** is currently presenting to us with persistent hematochezia. Our \textcolor{red}{gynecological examination} shows a large tumor in the area of the lower rectum that infiltrates the intestinal wall in the area of the lower rectum.
We informed Ms. ***** about the possible risks and the expected benefits of interstitial brachytherapy and she \textcolor{red}{agreed to the therapy}.
The following appointments have already been made for the patient:

06.08.2020 preparation for brachytherapy procedure.

10.08.2020 brachytherapy implantation.

...\\
\hline
\center{\textbf{Reference}} & ...

The patient underwent neoadjuvant radiochemotherapy for the rectal carcinoma she was diagnosed with in the above-mentioned stage at the Regensburg Clinic.
However, Ms. ***** refuses surgical resection of the rectal carcinoma \textcolor{red}{due to the risk of a possible colostomy.
The patient is afraid that the plaster used in the colostomy area would trigger a strong allergic reaction.}
We carried out a \textcolor{red}{clinical and sonographic examination} of the patient. This showed the previously described tumor, measuring approximately 2 cm to 3 cm, at 5 o'clock SSL.
Brachytherapy would in principle be technically feasible, but even with brachytherapy there is \textcolor{red}{a risk that a colostomy will be necessary  due to toxicity caused by the brachytherapy.}
For this reason, Ms. ***** is currently \textcolor{red}{opposed to this treatment option}, so we have referred her back to the Regensburg Clinic for re-evaluation of surgical resection of the known rectal carcinoma.
If the patient changes her mind, she can be re-presented at any time.

...\\
\hline
\end{tabular}
\end{small}
\caption{The LLM input, output, and reference from the original physician letter for Case $\#$9. The texts in red highlight the major differences between the fine-tuned LLaMA-3's output and the original physician letter.}
\label{table:Case9}
\end{table}

In Case $\#$9 displayed in Tab.\,\ref{table:Case9}, the generation of a physician letter for a female patient with recurrent rectal carcinoma was evaluated. The fine-tuned model's prediction and the original letter both emphasized the patient's refusal of surgical resection to treat the rectal carcinoma. However, the original letter provided more detailed information about the reason for her decision: due to the risk of a possible colostomy, the patient feared that the plaster used in the colostomy area would trigger a severe allergic reaction. The critical difference between the prediction and the original letter lies in the patient's decision regarding interstitial brachytherapy. The fine-tuned LLaMA-3 LLM predicted that the patient agreed to interstitial brachytherapy, and dates for the planned brachytherapy procedures were scheduled. In contrast, the original letter indicated that the patient refused the interstitial brachytherapy option due to the risk of toxicity, which also bears the risk of secondarily requiring a colostomy due to the high complication risk. Consequently, she was referred back to her original treatment center for re-evaluation of surgical resection. Additionally, the fine-tuned model inaccurately hallucinated a gynecological examination showing a large tumor infiltrating the intestinal wall in the lower rectum, which is not consistent with the provided local tumor stage of rcT2 in the input data. In reality, a clinical and sonographic examination was performed, revealing the previously described tumor measured approximately 2 cm to 3 cm, located at the 5 o'clock position in the subserosal layer (SSL). Due to these inconsistencies, the prediction received the lowest mean score of 2.5.

\section{Discussion}
This work demonstrates that a local LLM (LLaMA-3) model can be fine-tuned within the infrastructure of a hospital using institution-specific data to create a generative AI application for physician letter writing. We found that the locally fine-tuned model successfully learned the institution-specific style and content of the physician letters, which is exemplified in Fig.\,\ref{Fig:OutputExample2} and in the examples provided in the Supplements. This included the salutation and the signatures of the letter with the correct titles of the physicians, but also the sequence of information in the main text, the content elements of the letter, the style of writing as well as commonly used expressions. In stark contrast, a non-fine-tuned LLaMA model was not capable of producing any reasonable output for the related task of case summarization (Fig.\,\ref{Fig:OutputExample}). Hospitals possess a large amount of patient data that forms the ideal training corpus for developing institution-specific LLM-based applications. For this work, 14,479 physician letters could easily be downloaded and processed for local LLM fine-tuning. This wealth of data within hospitals currently can only be hardly tapped without local model training, because of data privacy regulations as well as data safety concerns. Local LLM fine-tuning and inference can avoid any sharing of data to AI hosting providers, increasing patient data safety as well as independence from centralized institutions. Decentralized training and local execution of LLMs could make health-care systems more resilient, because internet service providers (ISPs) as well as AI hosting companies can form single point of failures that could widely affect health-care services. Whereas de novo training of LLaMA-3-8b had been performed by Meta AI on 16,384 H100 80 GB GPUs requiring 1.3 million GPU hours \cite{dubey2024llama}, LLaMA-3-8b model fine-tuning with the QLoRA technique in this work was possible in 58 hours with a single 48Gb Nvidia RTX A6000 GPU on a hospital workstation. It is interesting to note this vast decrease in computational requirements for fine-tuning an LLM as compared to de novo training, which is enabled by LoRA (Low-rank adaptation) \cite{LORA} combined with quantization (i.e., QLoRA) \cite{dettmers2024qlora} and makes local development of specialized LLMs within hospitals feasible.

The LLM-generated physician letters overall received decent ratings by the five physician evaluators, especially in the category of practicality (i.e., benefit in practice, mean 3.44 out of 4). Therefore, we are further planning for a real clinical implementation of the developed letter generation model via an intranet web interface in the context of a prospective clinical trial.
However, several limitations of LLM-based physician letter generations must be considered. Since the model has been fine-tuned to generate physician letters based on the provided case-specific input information, it has limited capacity to generate content beyond the provided input data. The results of Case $\#$2 (Tab.\ref{table:Case2}) and Case $\#$9 (Tab.\ref{table:Case9}) revealed this limitation in the physician letter generation tasks. In the input information not all details of the conversation between the patient and the clinician were included. Consequently, the fine-tuned model is restricted in its ability to add such information, such as the reason for the refusal of surgical resection in Case $\#$9. Nevertheless, Case $\#$2 and Case $\#$9 show that the fine-tuned LLaMA-3 model has a certain ability to deduct the content beyond the input information, despite of inaccuracy. 
With more training data or extended input information, the fine-tuned model could show improved performance on such challenging cases. Nevertheless, it is mandatory for physicians to carefully review and correct the LLM-predicted letter in every patient case in a similar fashion to other automation tasks within radiation oncology \cite{huang2022deep,erdur2024deep,weissmann2023deep}. 
The results of the present evaluation indicate that this manual review is possible in $\leq$1 minute for most cases. Another limitation relates to privacy concerns when sharing or publishing LLM models fine-tuned on institutional data. In contrast to other tasks like auto-segmentation, where interinstitutional sharing of model weights has been proposed as a solution for privacy-preserving training on multicenter data \cite{huang2024multicenter}, it cannot be excluded that privacy-sensitive information could be extracted from the fine-tuned LLM.

While LLaMA-3 8b has no formal multilingual support \cite{dubey2024llama}, it is interesting to observe that the fine-tuned model in general showed good performance with the German physician letter task. This finding can be explained by the fact that LLaMA-3 8b nevertheless was pretrained on multilingual data. Moreover, the local fine-tuning was performed on a considerable amount of German physician letters for a large number of iterations. We only observed one potentially language related limitation regarding the date format. The date format in English, especially in the United States, is mm-dd-yyyy, while the date format in Germany is dd.mm.yyyy. In the evaluation, we found that the fine-tuned LLM in general could correctly handle the German date format but made mistakes in the presence of errors within the input data. For example, in the original medical record shown in Tab.\,\ref{table:10}, the doctor accidentally put the start time of the treatment at the end position inducing an error in the LLM’s physician letter prediction.

\begin{table}[h!]
\begin{small}
\centering
\begin{tabular}{|p{8.25cm}|p{8.25cm}|}
\hline
\textbf{Input with errors} & \textbf{Output of Fine-tuned Llama 3-8b} \\
\hline
\textcolor{red}{10.03.2014 - 04.03.2014}: Chemotherapy: CCNU (100 mg/m\(^2\) orally, day 1) Procarbacin (60 mg/m\(^2\) orally, days 8-21) & the patient received chemotherapy from \textcolor{red}{October 2014 to March 2014}, but it was discontinued due to severe side effects \\
\hline
\end{tabular}
\end{small}
\caption{An example with date-related input errors inducing a misinterpretation of the date format (mm-dd-yyyy vs. dd.mm.yyyy) within the model's output.}
\label{table:10}
\end{table}

At the time of writing, we did not find any prior studies reporting local fine-tuning of LLMs for institution-specific physician letter generation. However, several research papers \cite{tung2024comparison,ruinelli2024experiments,schoonbeek4835935completeness} have recently explored using general purpose LLMs like ChatGPT-4 with zero-shot prompting to automatically create physician letters \cite{guo2024qub} and patient case summaries \cite{barak2024harnessing}. Min Tung et al. \cite{tung2024comparison} used ChatGPT-4 to generate discharge letters in urologic patients. The authors performed zero-shot prompting of ChatGPT-4, while appending the case-specific medical record to the input prompt. The ChatGPT-4-generated discharge letters were subsequently compared against manually written letters created by junior physicians in a single-blinded fashion. Interestingly, GPT-4 created letters were superior to human-generated letters regarding information provision, while there was no significant difference in all other investigated domains including overall satisfaction of the blinded physician rater panel. Ruinelli et al. employed a similar strategy providing ChatGPT with patient-specific clinical notes and an input prompt to create discharge summaries in Italian for medical and surgical cases \cite{ruinelli2024experiments}. In addition, Schoonbeek used GPT-4 through an electronic health record system to create patient case summaries in Dutch language. Though GPT-4-generated patient summaries were less concise than those written by physicians, overall evaluation scores were equal and there even was a slight preference towards the LLM-created summaries (57\% vs. 43\%) with the ten physician raters \cite{schoonbeek4835935completeness}. Conversely, Guo et al. \cite{guo2024qub} used LLaMA-3-8b without fine-tuning to automatically create two specific sections of the medical discharge letter (``Brief Hospital Course” and ``Discharge Instructions”) . Similarly to the aforementioned approaches, the authors also designed a dedicated zero-shot prompt including the patient-specific medical information and achieved high NLP-evaluation metrics. All of these studies together with the observations from the present work suggest that LLMs have significant potential in supporting hospitals and clinicians with clinical documentation tasks and physician letter writing. However, despite the widespread use of OpenAI GPT-4 in most studies, its practical application in clinical settings with real patient data is often hindered or outright prohibited by data privacy regulations in many jurisdictions. Therefore, studies on open-source LLMs like LLaMA-3, which can be implemented within a hospital's IT infrastructure, are of particular relevance.

\section{Conclusion}

In the field of radiation oncology, the automatic generation of physician letters holds significant clinical value. Our study has demonstrated that base LLaMA models without fine-tuning are inadequate for generating physician letters effectively. However, the QLoRA algorithm offers an efficient method for fine-tuning LLaMA models, even with limited computational resources, while preserving data privacy. We have shown that the 8B LLaMA-3 model can be successfully fine-tuned on a 48 GB GPU using QLoRA. The fine-tuned model has effectively learned radiation oncology-specific information and can generate physician letters in an institution-specific style, thereby providing substantial practical value in assisting physicians with letter generation.

\end{document}